\documentclass{article}
\usepackage[utf8]{inputenc}

\title{Expert System for Diagnosis of Chest Diseases Using Neural Networks}

\author{Ismail Kayali\thanks{M.S.c Big Data Systems, Faculty Business and Management, National Research University Higher School of Economics. Moscow, Russia. E-mail: ikayali@edu.hse.ru}}

\date{Feb 14 2018}

\begin{document}

\maketitle

\begin{abstract}
This article represents one of the contemporary trends in the application of the latest methods of information and communication technology for medicine through an expert system helps the doctor to diagnose some chest diseases which is important because of the frequent spread of chest diseases nowadays in addition to the overlap symptoms of these diseases, which is difficult to right diagnose by doctors with several algorithms:
\textbf{Forward Chaining},\textbf{ Backward Chaining},\textbf{ Neural Network(Back Propagation)}

However, this system cannot replace the doctor function, but it can help the doctor to avoid wrong diagnosis and treatments. It can also be developed in such a way to help the novice doctors.

\end{abstract}

\section{Introduction}
Did you think beforehand when you were sick to ask your computer for medical advice? Is it reasonable to leave your computer making decisions about your health? Will we call our personal computers Dr. Computer?Well, you would probably do more in the near future.  because the ability of computer for store data and processed it becomes bigger in order to get useful information.

This system can be used by the domain experts themselves in order to retrieve
diagnoses and treatments quickly. It can also be used by novices who have recently
graduated from the medical school in order to get the best diagnosis for specific
symptoms and then generate the treatments of this diagnosis. In addition to those users, the system can be a very influential teaching tool in the faculty of medicine and health institutions.
The system assumes that users have some medical knowledge about diagnosis of
diseases and a good background in order to use personal computers.

Diagnostic rules will be inspired by the experts, doctors, specialized in this kind of diseases. The system performance can be increased by applying some machine learning methods on these rules and expertise. Moreover, the system can recommend treatments for the diseases in its specialism. These treatments are suggested based on many conditions and constraints related to the patient and the diagnosed disease.

Based on the above, the system has six major components: standard interfaces,
administration interfaces, knowledge-base editor, inference engine, explanation
subsystem, and management system. The first two components have been clarified
before. The knowledge-base editor allows the high-level users to add, remove and edit
rules in the knowledge-base. The inference engine is used to reason with the expert
knowledge, which is extracted from our experts and data specific to the particular
problem being solved. The inference engine is the most important component in the
system. The explanation subsystem allows the program to explain its reasoning to the
user. This component brings the life to the system! The last component, management
system, manages the users in the system. It defines the permissions for every user and routes the control when the user logs in. Moreover, all the permanent and temporary data are stored in a database which is designed to support all the system components.

Finally, the system has to support working over networks; either local network or
internet. This advantage will facilitate the access to the system from many experts and
knowledge engineers. Consequently, the knowledge-base can be expanded quickly to
cover many medical domains. This diversity in expertise will make the system more
robust and reliable.

\section{Applicability Statement}
Basic features will be included and simple inference
approaches will be used. Two inference algorithms, which are used in the rule-based systems, These algorithms are backward and
forward chaining algorithms. Moreover, providing explanation facilities to interpret the system behavior and output. After finishing the basic requirements, providing the system with some learning capabilities using Data Mining approaches and algorithms.

\section{Knowledge Representation}

The knowledge base contains the domain-specific knowledge required to solve the
problem. The knowledge base is created by the knowledge engineer, who conducts a
series of interviews with the expert and organises the knowledge in a form that can be directly used by the system. The knowledge engineer has to have the knowledge of
KBES technology and should know how to develop an expert system using a
development environment or an expert system development shell. It is not necessary that the knowledge engineer be proficient in the domain in which the expert system is being developed. But a general knowledge and familiarity with the key terms used in the domain is always desirable, since this will not only help in better understanding the domain knowledge but will also reduce the communication gap between the knowledge
engineer and the expert. Before deciding on the structure of the knowledge base, the
knowledge engineer should have a clear idea of different knowledge representation
schemes and the suitability of each under different circumstances. [1]
Developing an expert system involves tasks such as acquiring knowledge from an
acknowledged domain expert, documenting it and organizing it, generating a knowledge
net to check the relationships between different knowledge sources, checking for
consistency in the knowledge and finally transforming the knowledge net into a computer program using appropriate tools. The same piece of knowledge can be represented using more than one formal scheme, but with varying degrees of difficulty. The difficulty is not in the representation of the knowledge, but in its usage. The decision on selection of a scheme primarily depends on the type of application being built. Also, different knowledge representation schemes can be adopted for developing one application. [3]
The knowledge engineer has to decide which portion of the knowledge should be
represented in what form, depending on the nature of the knowledge and the efficiency of its use [5]. The most common methods of knowledge representation are:
\\1. Predicate logic
\\2. Production rules
\\3. Frames (objects) and semantic networks
\\4. Conventional programs
I have adopted the second mechanism, production rules, to represent the medical
expertise. This method is easy to formalize the medical knowledge especially in the
distinctive diagnosis which does not require a very deep knowledge in medicine.

\subsection{Production Rules}
Production rules are simple but powerful forms of knowledge representation providing
the flexibility of combining declarative and procedural representation for using them in a unified form. A production rule has a set of antecedents and a set of consequences. The antecedents specify a set of conditions and the consequences a set of actions.
A value assigned to a variable can be used to represent facts in the rules. This
representation is called a variable-operator-value triple. when the IF portion of a rule is satisfied by the facts stored in the context or fact established by a user input, the actions specified in the THEN portion are performed, and the rule is then said to be fired [2].
It is a common practice in the development of expert systems to logically divide the
rules into smaller rule bases and to control from a higher-level rule base which has
knowledge about the different rule bases in the knowledge base. If there are more than one rule bases, each of them should have separate working memories. The higher-level rule base having meta-rules will control the overall inference process and will have a global context. The inference engine of the expert system shell should properly handle different contexts for proper solution of the problem [6].

\subsection{Inference Mechanisms}
Inference mechanisms are control strategies or search techniques, which search through the knowledge base to arrive at decisions. The knowledge base is the state space and the inference mechanism is a search process. As expert systems predominantly process symbols, the inference process manipulates symbols by selection of rules, matching the symbols of facts and then firing the rules to establish new facts. This process is continued like a chain until a specified goal is arrived at.
In an expert system, the inference can be done in a number of ways. The two popular
methods of inference are backward chaining and forward chaining. Backward chaining is
a goal-driven process, whereas forward chaining is data-driven. The inference engine of our system uses backward chaining algorithm to infer facts.[4]

\subsection{Backward Chaining}
It is the goal-driven reasoning. In backward chaining, an expert system has the goal (a hypothetical solution) and the inference engine attempts to find the evidence to prove it.
First, the knowledge base is searched to find rules that might have the desired solution.
Such rules must have the goal in their THEN (action) parts. If such a rule is found and its
IF (condition) part matches data in the database, then the rule is fired and the goal is
proved. However, this is rarely the case.
The inference engine puts aside the rule it is working with (the rule is said to
stack) and sets up a new goal, a sub-goal, to prove the IF part of this rule. Then the knowledge base is searched again for rules that can prove the sub-goal. The inference engine repeats the process of stacking the rules until no rules are found in the knowledge base to prove the current sub-goal [7]. Then, the user is asked to provide a value for this goal and the inference engine goes back recursively.
There are three basic data structures used to implement the backward chaining
algorithm. They are the working memory (context), rule stack and goal stack. When any
one of the actions of inference process i.e. select, match and execute occurs, one or more of these data structures get modified [9].

\subsection{Forward Chaining}
It is the data-driven reasoning. The reasoning starts from the known data and proceeds forward with that data. Each time only the topmost rule is executed. When fired, the rule adds a new fact in the database. Any rule can be executed only once. The match-fire the cycle stops when no further rules can be fired [8].
Moreover, this method is used to gather information and then infer from it whatever
can be inferred. Hence, many rules may be executed that have nothing to do with the
established goal.

\subsection{Hybrid Inference}
The third inference strategy is the hybrid inference mechanism. It is a combination of the backward and forward chaining process. Backward chaining is suitable, if there are few goal states and many initial states. The sequence in which the data are requested depends on the flow of the inference process. In forward-chaining all the data that the user knows have to be given a prior and the system does not prompt for any data. If the data are
sufficient the goal may be reached; otherwise it may exit stating that the goal is
unreachable with the available data. Hence, the forward chaining process is suitable only when there are very few initial states and many goal states. In large expert systems, many initial and goal states can exist. Also, the user may not be able to give all the necessary data a prior, as he/she may not be able to visualize the flow of the inference process.
In this context, a hybrid inference mechanism seems to be suitable. It starts in the
forward chaining mode with an empty context by trying to establish the facts as they
appear in the rule base and then backward chains to prove or disprove them. The overall direction of the hybrid chaining process is forward with backward chaining being invoked as and when facts are to be established. The user does not give all the data a prior but gives data as and when prompted for [8].

\section{Project Plan}
The major work activities, which must be achieved in this project, are summarized below
in a hierarchical format in order to be readable. They are numbered according to the
Work Breakdown Structure (WBS) levels. The activities are as following:
\subsection{Writing Software Project Management Plan}
1. Preliminary Analysis of the System.
\\2. Understanding sections of the SPMP document.
\\3. Drawing the Block Diagram of the System.
\\4. Defining the Project Activities and Deliverable.
\\5. Drawing the activities and WBS diagrams.
\\6. Generating SPMP Document.

\subsection{Building a Prototype}
1. Specifying the system components
\\2. Specifying the Standard Part.
\\3. Specifying the Inference Engine.
\\4. Specifying the Knowledge-base.
\\5. Specifying the Knowledge-base Editor.
\\6. Specifying the Explanation System.
\\7. Specifying the Administrator Part.
\\8. Specifying the Management System.
\\9. Generating Business Requirement Specification Document.
\\10. Designing the system components.
\\11. Designing the Inference Engine.
\\12. Designing the Knowledge-base.
\\13. Designing the Knowledge-base Editor.
\\14. Designing the Explanation System.
\\15. Designing the Administrator Part.
\\16. Designing the Management System.
\\17. Generating Software Design Document (UML modeling)
\\18. Implementing the system components.
\\19. Implementing the Standard Part.
\\20. Implementing the Inference Engine.
\\21. Implementing the Knowledge-base.
\\22. Implementing the Knowledge-base Editor.
\\23. Implementing the Explanation System.
\\24. Implementing the Administrator Part.
\\25. Implementing the Management System.
\\26. Generating the Technical Report
\\27. Testing the system components
\\28. Testing the Standard Part.
\\29. Testing the Inference Engine.
\\30. Testing the Knowledge-base.
\\31. Testing the Knowledge-base Editor.
\\31. Testing the Explanation System.
\\32. Testing the Administrator Part.
\\33. Testing the Management System.
\\34. Generating the Testing Document.
\\35. Integrating all the system components.

\subsection{Building the Medical Dataset}
1. Defining the system specialism.
\\2. Collecting adequate information.
\\3. Manipulating the information and extracting rules.
\\4. Populating the knowledge-base with this information.

\subsection{Customizing the prototype}
1. Revising the specification document.
\\2. Revising the design document.
\\3. Revising the implementation.

\subsection{Testing the system}
1. Testing using Black Box Test methods.
\\2. Testing using White Box Test methods.
\\3. Integration Testing.
\\4. Acceptance Testing.

\subsection{Releasing the final software product}
1. Packaging the product.
\\2. Deployment the system on web server.

\section{Conclusion and Future Work}

The system aims to develop and produce highly interactive system for diagnosing chest
diseases and suggesting treatments. Therefore, the system can be used in clinics, medical
centers and hospitals. It enables the novices to work in the cases where the professionals
are absent. Moreover, the system can be very influential teaching tool in the faculty of
medicine and health institutes.
The experts can also use the system for generating solutions quickly since there are a
lot of similar cases can be diagnosed for one patient.Also, they can use the system to
archive the cases they encounter during their career.
Then, they use the data mining tool,supported by the system to discover new medical cases.
The ability of neural networks to classify and separate convergent data is able to diagnose chest diseases, despite the severe overlap between disease symptoms that have good results in disease diagnosis based on the medical preview.
We hope the future system will develop to include all possible chest diseases by securing a larger range of samples as they represent the biggest obstacle in artificial intelligence applications.

\end{document}